\pdfoutput=1

\documentclass[11pt]{article}

\usepackage[final]{acl}

\usepackage{times}
\usepackage{latexsym}
\usepackage{amsmath} 
\usepackage[T1]{fontenc}

\usepackage[utf8]{inputenc}

\usepackage{microtype}

\usepackage{inconsolata}

\usepackage{graphicx}

\usepackage{booktabs}
\usepackage{xspace}
\usepackage{multirow}
\newcommand{\dataset}[0]{\textsc{MMEvalPro}\xspace}

\title{\dataset: Calibrating Multimodal Benchmarks Towards Trustworthy and Efficient Evaluation}

\author{%
  Jinsheng Huang\textsuperscript{1,2,3}$^*$,
  Liang Chen\textsuperscript{1, 2}$^*$,
  Taian Guo\textsuperscript{1,2,3},
  Fu Zeng\textsuperscript{4},
  Yusheng Zhao\textsuperscript{1,2,3} \\
  \textbf{
  Bohan Wu\textsuperscript{1,2,3},
  Ye Yuan\textsuperscript{1,2,3},
  Haozhe Zhao\textsuperscript{1},
  Zhihui Guo\textsuperscript{5},
  Yichi Zhang\textsuperscript{1},
  Jingyang Yuan\textsuperscript{1,2,3}} \\
  \textbf{
  Wei Ju\textsuperscript{1,2,3},
  Luchen Liu\textsuperscript{1,2,3},
  Tianyu Liu\textsuperscript{6},
  Baobao Chang\textsuperscript{1, 2}$^{\dagger}$,
  Ming Zhang\textsuperscript{1, 2, 3}$^{\dagger}$} \\
  \textsuperscript{1}National Key Laboratory for Multimedia Information Processing, Peking University \\
  \textsuperscript{2}School of Computer Science, Peking University,  
  \textsuperscript{3}PKU-Anker LLM Lab
  \\
  \textsuperscript{4}Chinese Academy of Medical Sciences,
  \textsuperscript{5}CUHK,
  \textsuperscript{6}Alibaba Group\\
  \texttt{hjs@stu.pku.edu.cn}\quad\texttt{leo.liang.chen@outlook.com}\quad
  \texttt{\{chbb,mzhang\_cs\}@pku.edu.cn}\\
  \href{https://mmevalpro.github.io}{\large{\color{magenta}{\textit{https://mmevalpro.github.io}}}}
  \\
}

\begin{document}

\maketitle

\def\thefootnote{$^*$}\footnotetext{Equal contribution. $^\dagger$Corresponding authors.}
\def\thefootnote{\arabic{footnote}}

\begin{abstract}

Large Multimodal Models (LMMs) exhibit impressive cross-modal understanding and reasoning abilities, often assessed through multiple-choice questions (MCQs) that include an image, a question, and several options. However, many benchmarks used for such evaluations suffer from systematic biases. Remarkably, Large Language Models (LLMs) without any visual perception capabilities achieve non-trivial performance, undermining the credibility of these evaluations. To address this issue while maintaining the efficiency of MCQ evaluations, we propose \dataset, a benchmark designed to avoid Type-I errors through a trilogy evaluation pipeline and more rigorous metrics. For each original question from existing benchmarks, human annotators augment it by creating one perception question and one knowledge anchor question through a meticulous annotation process. \dataset comprises $2,138$ question triplets, totaling $6,414$ distinct questions. Two-thirds of these questions are manually labeled by human experts, while the rest are sourced from existing benchmarks (MMMU, ScienceQA, and MathVista). Compared with the existing benchmarks, our experiments with the latest LLMs and LMMs demonstrate that \dataset is \textbf{more challenging} (the best LMM lags behind human performance by $31.73\%$, compared to an average gap of $8.03\%$ in previous benchmarks) and \textbf{more trustworthy} (the best LLM trails the best LMM by $23.09\%$, whereas the gap for previous benchmarks is just $14.64\%$). Our in-depth analysis explains the reason for the large performance gap and justifies the trustworthiness of evaluation, underscoring its significant potential for advancing future research.

\end{abstract}

\section{Introduction}

\begin{figure}[!h]
    \centering
    \includegraphics[width=\linewidth]{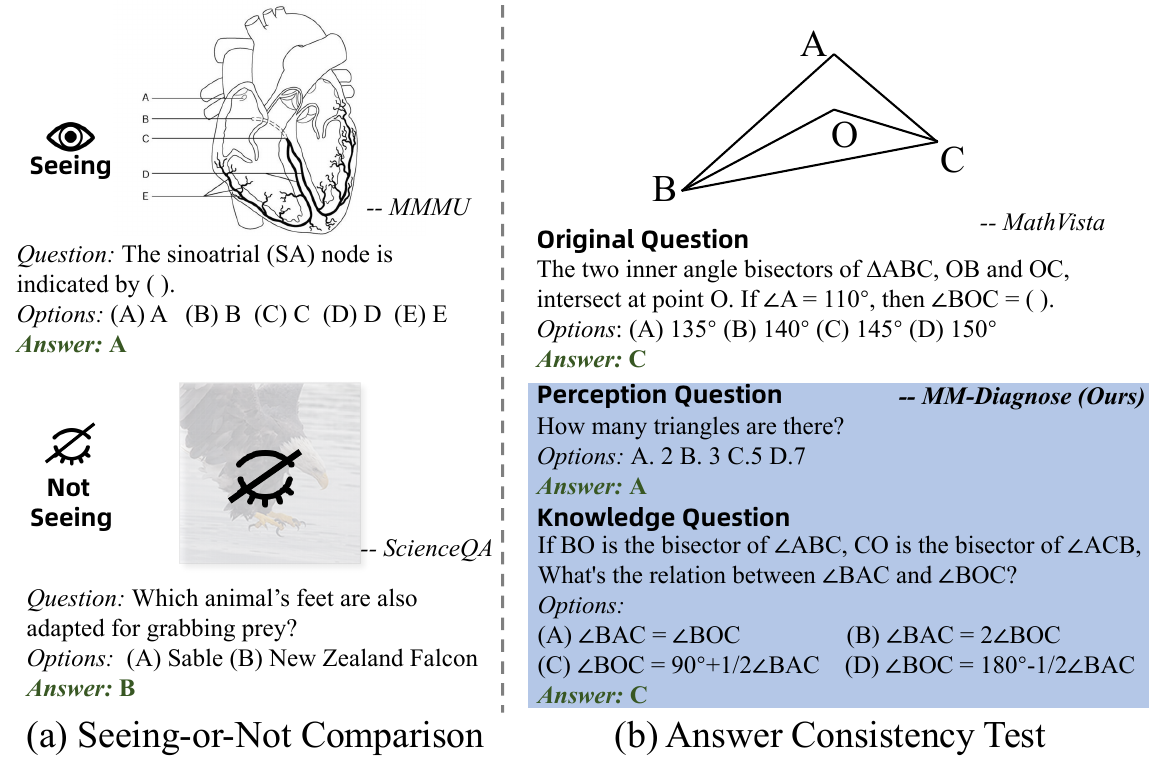}
    \caption{Examples of the probing experiments.}
    \label{fig:probing-examples}
\end{figure}
\begin{figure}[!h]
    \centering
    \includegraphics[width=\linewidth]{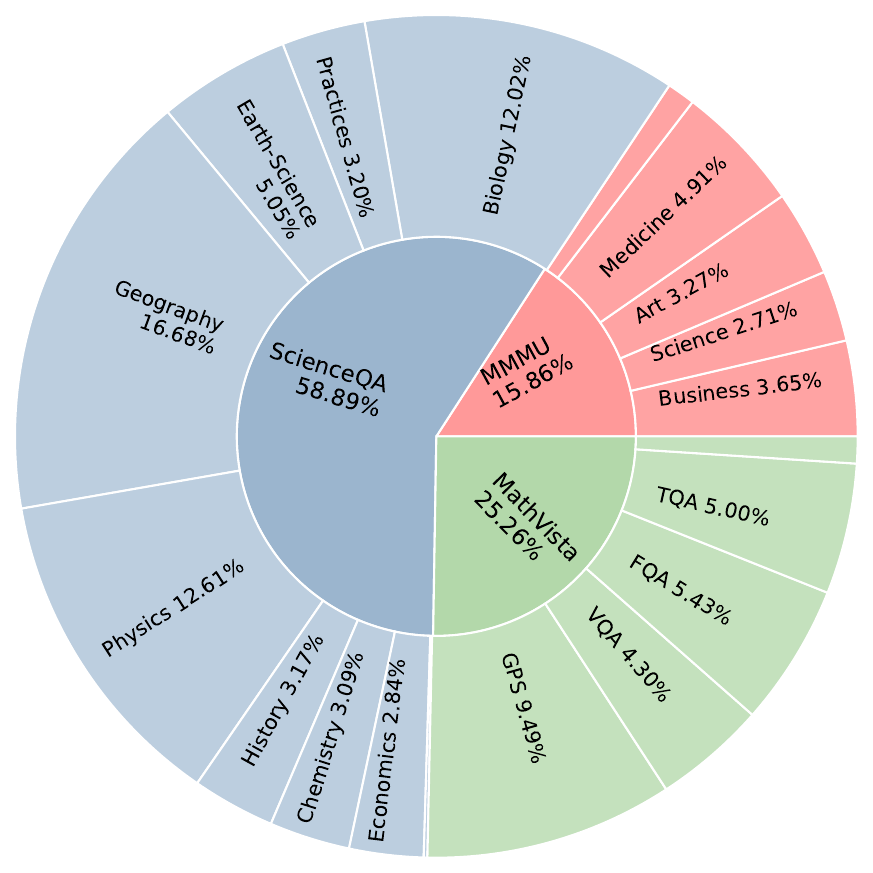}
    \caption{Topic distribution  of \dataset's data.}
    \label{fig:dataset-dist}
    \vspace{-10pt}
\end{figure}

\begin{figure*}[h]
\centering
\includegraphics[width=\textwidth]{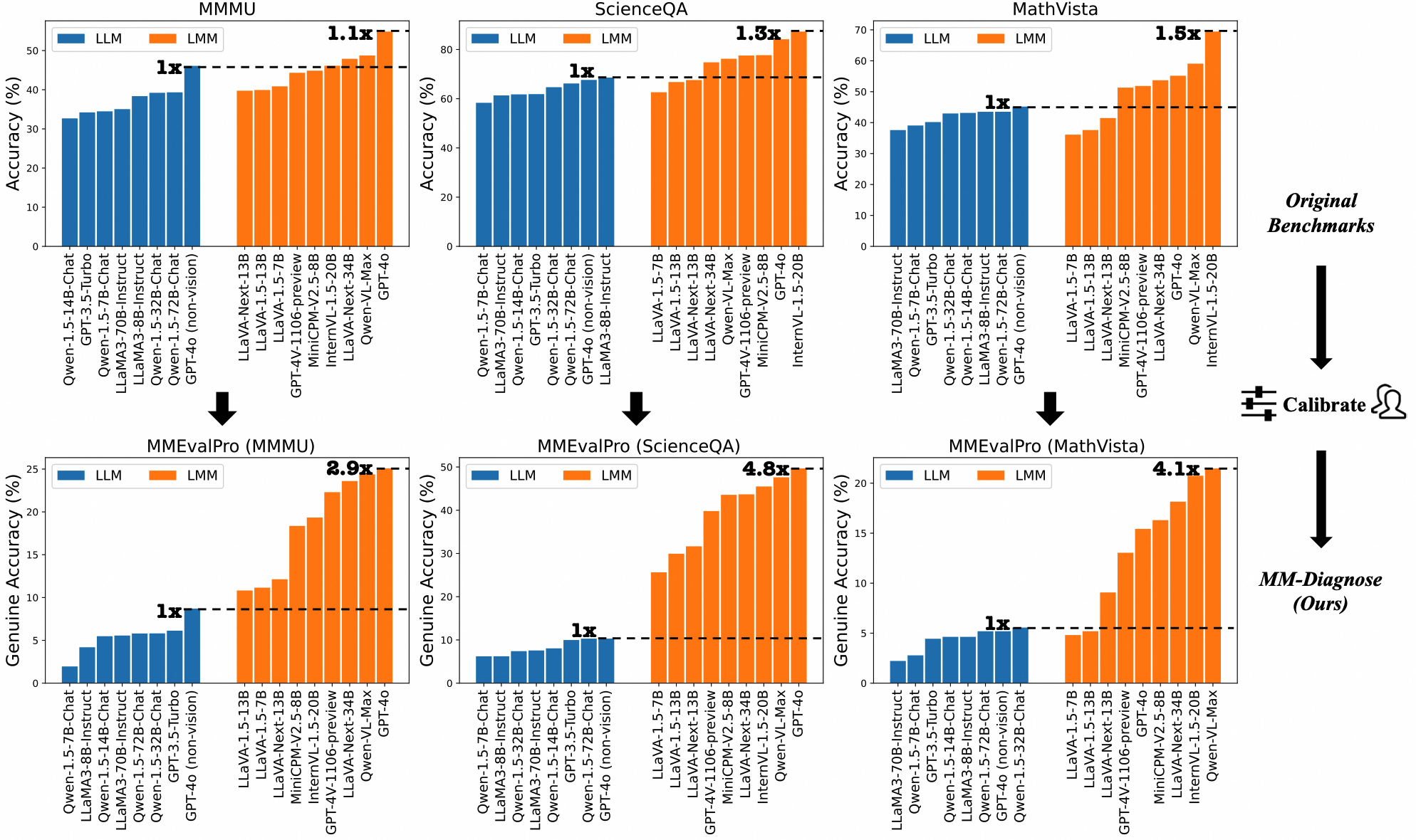}
\caption{LLMs and LMMs' performance comparison between original multimodal benchmarks and \dataset. Performance gap between LLM and LMM is much clearer in \dataset. }
\label{fig:see-or-not-bar}
\vspace{-5pt}
\end{figure*}

Ever since the birth of standardized testing, the credibility of its conclusions has been a significant concern. The same problem goes for the evaluation of recently popular Large Multimodal Models (LMMs) such as GPT4-o~\citep{gpt4o}, Gemini-1.5~\citep{geminiteam2024gemini}, Qwen-VL~\citep{QwenVL} and LLaVA~\citep{liu2023llava}. One classic composition of such an evaluation is the multiple-choice question (MCQ), which includes an image, a question, possible choices, and an answer. This form of evaluation has higher usability compared to other evaluation methods, such as text-based evaluation and human evaluation~\citep{chen2024next}. Many benchmarks~\citep{fu2023mme,liu2023mmbench,lu2024mathvista,lu2022scienceqa,yue2023mmmu,zhang2024cmmmu,chen2023endtoend,chen2024pcabench,Kembhavi2016ADI,shen2024measuring} designed for multimodal foundation models include a large portion of MCQs and are widely adopted in testing from basic to most advanced multimodal models~\citep{gpt4o,QwenVL,liu2023llava}, with some models reaching or even surpassing human scores on certain benchmarks~\citep{lu2022scienceqa,Kembhavi2016ADI}. However, it is important to consider whether these evaluation results truly reflect the absolute capabilities of the models, especially when we are comparing them with human beings in the pursuit of AGI. In fact, several work such as PCA-Bench~\citep{chen2024pcabench}, MMStar~\citep{mmstar} and MathVerse~\citep{zhang2024mathverse} have pointed out that the multimodal MCQ evaluation has intrinsic bias, which provides Large Language Models (LLMs) with shortcut to hack the question. FastV~\citep{chen2024imageworth12tokens} finds that LMM could even achieve better performance on some benchmarks with only partial visual tokens. In this paper, we mainly study three popular multimodal benchmarks MMMU~\citep{yue2023mmmu}, ScienceQA~\citep{lu2022scienceqa} and MathVista~\citep{lu2024mathvista}.

In our preliminary Seeing-or-Not Comparison experiment, we found that LLMs could achieve high scores without processing the visual data, attributed to possible data leakage, visual information problems being not related to answering the question or simply guessing the answer. Notably, the average performance gap between best LLM and best LMM is just 14.64\%, which is even smaller than the gap within LMM themselves, revealing the unreliable problem of such evaluations.

Our further Answer Consistency Test into multiple choice questions (MCQ) reveals a prevalent Type-I Error in such evaluation's conclusion, where models could output correct answers without actual comprehension. For example, the model could calculate the degree for a particular angle, but could not recognize the correct angle's name in the figure, which is a prerequisite to compute the degree.

To this end, we propose \dataset to truthfully reflect the true multimodal capabilities of tested models and keep the simplicity of MCQ evaluation. We achieve this by augmenting the original MCQ with prerequisite perception and knowledge questions. We propose Genuine Accuracy as the main metric, which depends on whether the model answers the triplet questions concurrently. Overall, in \dataset we annotate $2,138$ question triplets, originating from MMMU, ScienceQA, and MathVista, resulting in $6,414$ individual questions. 

We carry out experiments and analyses involving $17$ different models and human experts. The findings indicate that \dataset offers a more precise reflection of the tested LMMs' capabilities and poses a more demanding challenge. \dataset is more trustworthy than base benchmarks as the best LLM trails the best LMM by $23.09\%$ whereas the gap for previous benchmarks is just $14.64\%$. Significantly, even the most advanced models, including GPT-4o and Qwen-VL-Max, lag considerably behind human performance, with a notable gap of over $30\%$ in Genuine Accuracy (only $8.03\%$ for the base benchmarks). Our investigation into the factors contributing to the consistency gap illuminates the existing disparities and supports the evaluation credibility of \dataset, supplying insights for future research endeavors.

\section{Probing the Credibility of Multimodal Benchmarks}
The fundamental assumption of any benchmark is that models achieving higher scores possess superior capabilities. In this section, we question the credibility of such an assumption for existing multimodal benchmarks. We find that the existing benchmarks are not trustworthy enough in either relative or absolute perspectives, which is concluded from two probing experiments: Seeing-or-Not Comparison and Answer Consistency Test. The processes are illustrated in Figure~\ref{fig:probing-examples}. We test the MCQ evaluation in three multimodal benchmarks across various domains including MMMU, ScienceQA-Image and MathVista. We provide the detail of dataset statistics in section~\ref{sec:data-source}, model inference hyper-parameters and prompts in Appendix-\ref{app: Experiment Setup}.


\subsection{Seeing-or-Not Comparison}

As shown in Figure~\ref{fig:probing-examples}-(a), we prepare two data versions for each benchmark: ``Seeing'' (with image, question, options, and answer) and ``Not-Seeing'' (without image). We test leading LMMs on ``Seeing'' data and non-vision LLMs on ``Not-Seeing'' data, then compare the results. The outcomes are shown in the first row of Figure~\ref{fig:see-or-not-bar}.

The figure indicates that the performance gap between LMMs and LLMs is significantly narrower than anticipated. Intuitively, one might assume that LLMs, which is unable to process visual information, would perform considerably worse on multimodal benchmarks. In fact, if we compare the scores between the best-performing LLM and LMM, we observe that for MMMU, the best LMM's performance is only 1.1 times that of the best LLM. This performance gap is even smaller than the variability observed within LMMs. Similar results go for the other tested benchmarks. It's more surprising that LLMs sometimes outperform their vision-enabled counterparts (GPT4-o without vision ability outperforms the LLaVA-1.5 series according to Figure~\ref{fig:see-or-not-bar}) and there is not an apparent performance boundary between the two kinds of models. These results suggest that those benchmarks do not accurately reflect the true multimodal understanding capabilities of the tested models. The reason for this phenomenon is three-fold:

\textbf{1. Image is not needed:} Some benchmark questions can be answered solely through textual information, making visual input unnecessary. This diminishes the advantage of vision-enabled models (LMMs). For instance, as shown in the ScienceQA question in Figure~\ref{fig:probing-examples}, the knowledge that a falcon uses its feet to capture prey is common and does not require an image for verification. 

\textbf{2. Data leakage:} During training, LLMs may inadvertently encounter similar questions or datasets, leading to unfair advantages. The existing benchmarks often derive questions from textbooks~\citep{lu2022scienceqa,yue2023mmmu}, online education resources, and research papers~\citep{lu2024mathvista}, which are also sources for training datasets~\citep{touvron2023llama, clement2019arxiv}. 

\textbf{3. Educated guessing:} LLMs are trained on extensive text datasets, enabling them to make educated guesses even without visual information, which narrows the performance gap with LMMs.

\paragraph{Quantitative Analysis}

\begin{table}[!t]
\centering
\resizebox{0.5\columnwidth}{!}{
\begin{tabular}{lc}
 \toprule
 {\textbf{Dataset}} & {\textbf{Proportion}} \\
 \midrule
  MMMU & 2.97\% \\
  ScienceQA & 43.08\%\\
  MathVista & 5.37\%\\
 \bottomrule
 \end{tabular}}
 \captionof{table}{Proportions of “Image is not needed” samples in the original datasets}
 \vspace{-10pt}
 \label{tab:quant_ana}
\end{table}

We analysed the proportion of samples that can be answered correctly due to “Image is not needed” in the original datasets and the results are shown in the Table~\ref{tab:quant_ana}. As the results shown, the 43.08\% proportion in ScienceQA is the highest, which is also reflected in the "Not-Seeing" experiments. From analysis, we can concluded that “Image is not needed” samples significantly affected the credibility of previous benchmarks.


These factors collectively result in the unexpectedly narrow performance gap between LLMs and LMMs on multimodal benchmarks. To create a fair multimodal evaluation benchmark: (1) Ensure questions are intrinsically tied to the image details, making visual information essential for deriving the answer. 
(2) Prevent contamination of training data to ensure models are reasoning through problems instead of recalling memorized answers. 
(3) Design questions and answers to minimize the likelihood of the model making accurate guesses.

\subsection{Answer Consistency Test}
To determine whether a model truly understands a question or is simply "hacking" the answer, we simulate the human problem-solving process by creating "anchor questions" that must be answered before the main question. When humans tackle multimodal reasoning questions, they typically follow two key steps: (1) identifying relevant visual clues in the image, and (2) applying their knowledge to reason through the problem before arriving at the final answer. Omitting either step usually leads to an incorrect answer. Similarly, we create a perception anchor question and a knowledge anchor question related to the original question. If a model can answer both anchor questions and the final question, it demonstrates genuine comprehension and reasoning, rather than mere guessing.

An example is shown in Figure~\ref{fig:probing-examples}-(b), we set a perception question and a knowledge question to the MCQ from MathVista. For human examinees, the perception question and knowledge question are easier to answer than the original since they are the prerequisites for the original one in the solution path. We found that even the most advanced LMM such as GPT4o struggles at answering all related questions even if it answers the original question correctly, which is easy for human experts. A detailed case analysis is in Figure~\ref{fig:casestudy1}.


The phenomenon raises another concern for multiple-choice question (MCQ)-based multimodal evaluation benchmarks: correctly answering a question does not necessarily indicate that the model genuinely knows how to derive the final answer. A direct solution to accurately diagnose whether the model truly has the capability to solve the question is to let humans evaluate the model's reasoning process, as done in previous works like MathVista~\citep{lu2024mathvista}. However, human evaluation is labor-intensive and not reproducible, which complicates the broader application of the method.

Several related works use advanced LLMs like GPT-4 to replace humans in evaluating the reasoning process of LMMs~\citep{chen2024pcabench,zhang2024mathverse}. These methods show a strong correlation with human judgments in corresponding tests, but they result in unstable evaluation due to updates of proprietary models and inevitable API costs. Research in using LLMs as evaluators~\citep{Wang2023LargeLM, chen2024pcabench} also finds systematic bias and a significant gap between open-source and proprietary models in terms of evaluation agreement with human experts. These drawbacks highlight the need for an economical, easy-to-use, and calibrated method for multimodal models.



\section{\dataset: Calibrating Multimodal Evaluation}

In this section, we delve into the detailed process of constructing the \dataset benchmark dataset and elucidate our methodologies for ensuring high-quality evaluation standards. Through these efforts, \dataset sets a new paradigm in the assessment of multimodal models, aiming to foster both accuracy and efficiency in multimodal evaluation. 

\subsection{Data Source}
\label{sec:data-source}

To enhance the diversity of our benchmark data, \dataset integrates content from three prominent multimodal benchmarks: MMMU, ScienceQA, and MathVista. These benchmarks span educational levels from junior high to undergraduate and cover various subjects. The details of the source datasets are in Appendix~\ref{app:data source}. Considering annotator expertise and budget, we selected $328$ questions from MMMU(dev), $1,200$ from ScienceQA-Image, and all $540$ from MathVista, totaling $2,138$ distinct multimodal MCQs.



\begin{figure*}[!t]
\centering
\includegraphics[width=1\textwidth]{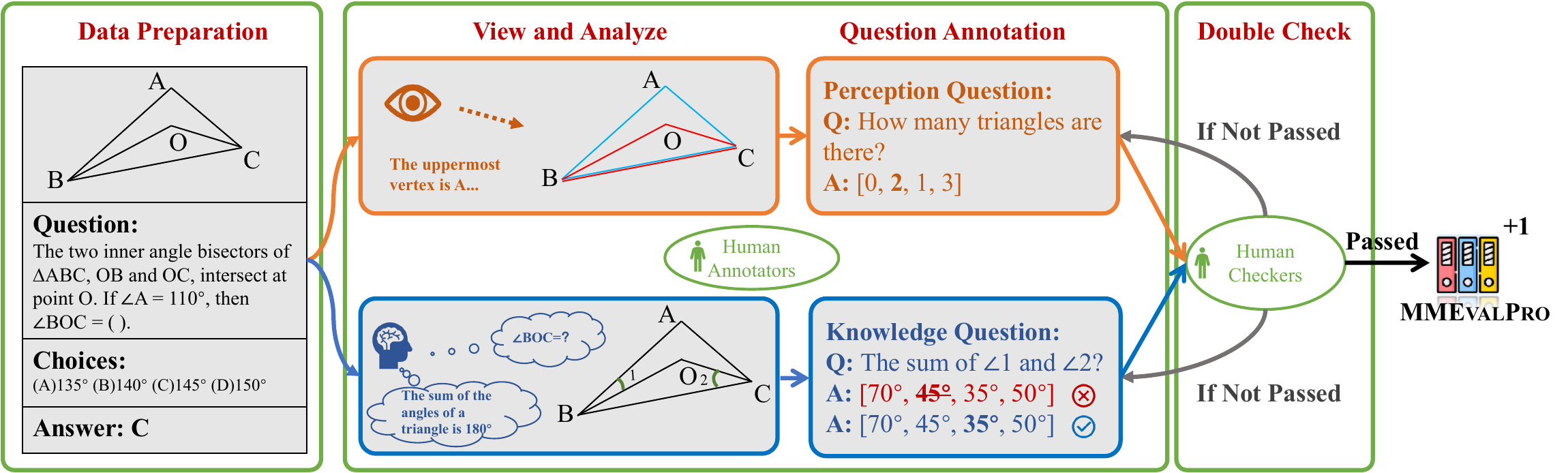}
\caption{Annotation pipeline for \dataset.}
\label{fig:mmd-pipelines}
\vspace{-5pt}
\end{figure*}

\subsection{Annotation Pipeline}

As illustrated in Figure~\ref{fig:mmd-pipelines}, we design the following pipeline for \dataset to generate question triplets. The triplet consists of an original question, a perception question, and a knowledge question.

\textbf{1) Data Preparation: } Annotators begin by thoroughly reviewing the original question to ensure a deep understanding of concepts and solutions.

\textbf{2) View and Analyse: } Annotators are tasked with extracting crucial visual information and the logical framework implicit in the original problem, paving the way for the creation of nuanced perception and knowledge questions.

\textbf{3) Question Annotation: } Building on the insights gathered, annotators then proceed to enrich the original question by formulating corresponding perception and knowledge questions, thereby expanding the scope of the evaluation.

\textbf{4) Double Check: } To maintain the integrity of the \dataset dataset, each annotated question triplet undergoes a rigorous verification process. Two independent checkers, who are not part of the annotation team, review each triplet for any errors or logical inconsistencies. Any issues identified prompt a re-annotation of the affected questions.

We provide the annotator guide in the supplement material. The final distribution of \dataset is shown in Figure~\ref{fig:dataset-dist}. We list the key statistics of the benchmark in Table~\ref{tab:statistics} from Appendix~\ref{sec:data-source}, the annotation guidelines in Appendix~\ref{app:annotation-guideline} and ~\ref{app:check-guideline}.




\subsection{Evaluation Metrics}

We propose \textbf{Genuine Accuracy (GA)} as the primary metric for \dataset. GA equals 1 only if the model correctly answers the original question and the corresponding perception and knowledge prerequisite questions simultaneously. The second metric is the \textbf{Average Accuracy (AA)}, which computes the average accuracy of all questions. 

\label{sec:metric}
\dataset can also be viewed as a multi-view evaluation process, where we naturally derive the \textbf{Perception Accuracy (PA)} score and the \textbf{Knowledge Accuracy (KA)} score by computing the average accuracy for the perception and knowledge anchor questions, respectively. 

The \textbf{Consistency Gap (CG)} is measured by subtracting the Genuine Accuracy from the accuracy of the original question. This metric reflects the proportion of instances where the model correctly answers the original question but fails on more in-depth perception and knowledge questions, leading to inconsistency in its answers. To gain deeper insights into the answer consistency and generalization capabilities of the tested models, we define \textbf{Perception Consistency (PC)} and \textbf{Knowledge Consistency (KC)} as the conditional probabilities $P(\text{Perception}=1 \mid \text{Origin}=1)$ and  $P(\text{Knowledge}=1 \mid \text{Origin}=1
)$, where ( $=1$ ) indicates correctly answering the corresponding question in a question triplet. PC and KC together reveal the answer consistency of the tested model.

\subsection{Quality Evaluation}

To ensure the precision and consistency of our annotations, all question triplets are annotated by graduate students. For specialized college-level subjects in the MMMU sub-set, we hire students in the corresponding major to ensure annotation accuracy. All annotators are required to thoroughly familiarize themselves with the guide before commencing their annotation tasks. During the annotation process, we assign a minimum of three annotators for one question triplet. These foundations ensure the quality of the new annotated $4,276$ questions and there are only $47$ questions that go through more than two double-check circles. Consensus was reached on all examples in the end.

\section{Experiments}

In this section, we evaluate a wide range of models on our \dataset benchmark. We first introduce the evaluation setup, and then present the quantitative results for both open-source and closed-source models. Finally we
investigate the answer consistency gap among LLMs, LMMs and human experts with fine-grained analysis.

\subsection{Setup}

We test multiple LLMs and LMMs on the original benchmarks and \dataset. To streamline the evaluation process, all questions are converted into a multiple-choice format, prompting the models to directly provide the answers. This approach simplifies answer matching process and eliminates the need for external models like ChatGPT, which is commonly used in previous studies such as~\citep{lu2024mathvista,chen2024pcabench,zhang2024mathverse}. In line with prior research, we incorporate an in-context demonstration for LLMs to standardize the output format. This is important as we have observed that different LLMs tend to produce varying response formats when not provided with an example. The detailed prompt and demonstration template are listed in Appendix~\ref{app: Experiment Setup}. For the original benchmarks, we report the average accuracy on the MCQ questions the same as we sampled for creating \dataset for fair comparison. For \dataset, we report both the Genuine Accuracy and Average Accuracy.

\subsection{Evaluated Models}

In our evaluation, we evaluate a variety of both LLMs and LMMs. For LLMs, we implement a 1-Shot setting, where a single demonstration is utilized to guide the output format. Among the LLMs assessed are some of the most advanced open-source models, including four versions of Qwen-1.5-Chat~\citep{bai2023qwen} with sizes spanning from 7B to 72B, as well as the LLaMA3-Instruct~\citep{touvron2023llama} series (8B and 70B). We also tested API-only models, such as GPT-3.5-Turbo~\citep{chatgpt} and GPT4-o~\citep{gpt4o}, which rely solely on language. On the other hand, LMMs are evaluated in a zero-shot manner due to their ability to follow instructions to produce valid answer choices. In the open-source category, we tested the LLaVA-1.5~\citep{liu2023llava15} and LLaVA-Next~\citep{liu2024llavanext} series, which include models of 7B, 13B, and 34B in size. We also evaluated two of the most cutting-edge LMMs, MiniCPM-V2.5-LLaMA3-8B~\citep{viscpm} and InternVL-1.5-Chat-20B~\citep{chen2023internvl}, both of which have recently been made available to the public. For proprietary models, we tested GPT-4V-1106-preview~\citep{gpt4v}, the latest GPT-4o~\citep{gpt4o}, and Qwen-VL-Max~\citep{QwenVL}. We provide the prompts for LLMs and the hyperparameters used for LMMs in Appendix~\ref{app: Experiment Setup}.

\begin{table*}[!t]
    \caption{Main experiments result. For average, we report the macro average scores (the mean of different domains). Highest score is marked \textbf{bold} and the second highest is \underline{underlined}. Best LMM still lags behind human with a substantial gap ($-31.73\%$ average {\color{red}{GA}}) on \dataset.}
    \centering
    \resizebox{2.0\columnwidth}{!}{
    \begin{tabular}{lc|cccc|cccc}
    \toprule
    \multirow{3}{*}{Model} & \multirow{3}{*}{Open?} & \multicolumn{4}{c|}{\textbf{Original Source}} & \multicolumn{4}{c}{\textbf{\dataset}} \\
    ~& ~& MMMU & ScienceQA &  MathVista & Average  &  MMMU & ScienceQA & MathVista & Average \\
    ~& ~& \multicolumn{4}{c|}{\textit{(Average Accuracy)}} & \multicolumn{4}{c}{\textit{({\color{red}{\textbf{Genuine Accuracy}}} / Average Accuracy)}} \\
    \midrule
    \multicolumn{10}{c}{\textit{Large Language Models (1-Shot) }}\\
    \midrule
    Qwen-1.5-7B-Chat & Yes & 34.48\% & 58.32\% & 39.07\% & 43.96\% & 1.95\% / 30.30\% & 6.21\% / 41.08\%  & 2.78\% / 29.19\% & 3.65\% / 33.52\%\\
    Qwen-1.5-14B-Chat & Yes &  32.68\% & 61.75\% & 43.14\% & 45.86\% & 5.48\% / 35.05\% & 8.05\% / 43.74\% & 4.63\% / 35.74\% & 6.05\% / 38.18\%\\
    Qwen-1.5-32B-Chat & Yes & 39.21\% & 64.66\% & 42.96\%& 48.94\% & 5.81\% / 39.56\% & 7.40\% / 45.31\% & 5.56\% / 37.53\% & 6.26\% / 40.80\% \\
    Qwen-1.5-72B-Chat & Yes & 39.33\% & 66.18\% & 43.52\% & 49.68\% & 5.80\% / 38.06\% & 10.30\% / 41.09\% & 5.19\% / 36.98\% & 7.10\% / 38.71\%\\
    LLaMA3-8B-Instruct & Yes & 38.36\% & 68.57\% & 43.51\%& 50.15\% & 4.19\% / 34.09\% & 6.22\% / 45.24\% & 4.63\% / 34.32\% & 5.01\% / 37.88\%\\
    LLaMA3-70B-Instruct & Yes & 35.56\% & 63.47\% & 37.59\% & 45.54\% & 4.84\% / 34.94\% & 8.05\% / 44.06\% & 2.59\% / 31.85\% & 5.16\% / 36.95\%\\
    GPT-3.5-Turbo & No & 34.21\% & 61.85\% & 40.18\%& 45.41\% & 6.13\% / 38.28\% & 9.98\% / 45.87\% & 4.44\% / 33.58\% & 6.85\% / 39.24\%\\
    GPT-4o (non-vision) &  No & 46.09\%  & 67.62\% & 45.19\%&52.97\% & 8.71\% / 39.90\% & 10.30\% / 48.18\% & 5.19\% / 37.78\% & 8.07\% / 41.95\%\\
    \midrule
    \multicolumn{10}{c}{\textit{Large Multimodal Models (Zero-Shot)}}\\
    \midrule
    LLaVA-1.5-Vicuna-7B  &Yes & 40.86\% &62.61\% &36.11\%& 46.53\% & 11.15\% / 43.06\% & 25.64\% / 58.33\% & 4.81\% / 37.10\% & 13.87\% / 46.16\% \\
    LLaVA-1.5-Vicuna-13B  &Yes & 39.92\% & 66.76\% &37.59\%& 48.09\% & 10.82\% / 43.28\% & 29.94\% / 63.20\% & 5.19\% / 36.98\% & 15.32\% / 47.82\%\\
    LLaVA-Next-Vicuna-13B  &Yes & 39.75\% & 67.57\% & 41.48\%& 49.60\% & 12.13\% / 47.86\% &31.65\% / 64.88\%  & 9.07\% / 39.26\% & 17.62\% / 50.67\%\\
    LLaVA-Next-Hermes-Yi-34B  &Yes & 47.89\% & 74.77\% & 53.70\%& 58.79\% & 23.60\% / 59.34\% & 43.67\% / 73.35\% & 18.15\% / 54.75\% & 28.47\% / 62.48\%\\
    MiniCPM-V2.5-LLaMA3-8B & Yes & 44.86\% & 77.68\% & 51.32\%& 57.95\% &18.36\% / 54.86\%& 43.56\% / 73.96\% & 16.30\% / 52.28\% & 26.07\% / 60.37\% \\
    
    InternVL-1.5-Chat-20B & Yes & 46.12\% & \textbf{87.27\%} & \textbf{69.44\%}& \textbf{67.61\%} & 19.34\% / 55.63\%  &45.49\% / \underline{76.25\%}  & \underline{20.74\%} / \underline{57.78\%} & 28.52\% / 63.22\% \\

    GPT-4V-1106-preview  &No & 44.32\% & 77.54\% & 51.85\%& 57.90\% & 22.30\% / 57.95\% & 39.81\% / 71.34\% & 13.04\% / 48.48\% & 25.05\% / 59.26\% \\
    GPT-4o  &  No &\textbf{54.85\%} & \underline{84.13\%} & 55.16\% & \underline{64.71\%} & \textbf{25.08\%} / \textbf{60.29\%} & \textbf{49.68\%} / \textbf{76.86\%} & 15.43\% / 52.63\% & \underline{30.06\%} / \underline{63.26\%}  \\
    
    Qwen-VL-Max     & No & \underline{48.75\%} & 76.22\% & \underline{59.07\%} & 61.35\% & \underline{24.38\%} / \underline{59.45\%} & \underline{47.61\%} / 75.02\% & \textbf{21.48\%} / \textbf{59.19\%} & \textbf{31.16\%} / \textbf{64.55\%} \\

    \midrule
    Random Guess& - & 26.04\% & 35.53\% & 32.41\%& 31.33\% & 1.94\% / 28.60\% & 2.36\% / 29.26\% & 3.32\%/ 29.14\% & 2.54\% / 29.01\%\\
    Human (Graduate Student)  & - & 49.21\% & 76.92\% & 92.08\% &72.74\%& 38.10\% / 66.67\% & 64.42\% / 81.09\% & 86.14\% / 93.07\% &62.89\% / 80.28\%\\
    \bottomrule
\end{tabular}}
    \label{tab:main-exp}
    \vspace{-4pt}
\end{table*}

\begin{table*}[!t]
    \caption{Fine-grained scores on \dataset. \textbf{CG:} Consistency Gap,
    \textbf{PA:} Perception Accuracy,
    \textbf{KA:} Knowledge Accuracy,
    \textbf{PC:} Perception Consistency,
    \textbf{KC:} Knowledge Consistency}
    \centering
    \resizebox{2.0\columnwidth}{!}{
    \begin{tabular}{lcc|ccccc|ccccc|ccccc}
    \toprule
    \multirow{2}{*}{Model}&\multirow{2}{*}{Type} &\multirow{2}{*}{Open?} & \multicolumn{5}{c|}{\dataset-MMMU} & \multicolumn{5}{c|}{\dataset-ScienceQA} & \multicolumn{5}{c}{\dataset-MathVista} \\
    ~&~&~& CG $\downarrow$ & PA $\uparrow$ & KA $\uparrow$ & PC $\uparrow$ & KC $\uparrow$ & CG$\downarrow$ & PA $\uparrow$ & KA $\uparrow$ & PC $\uparrow$ & KC $\uparrow$ &CG$\downarrow$ & PA $\uparrow$ & KA $\uparrow$& PC $\uparrow$ & KC $\uparrow$   \\
    \midrule
    Qwen-1.5-72b-Chat &LLM & Yes & 33.53\% & 32.58\% & 41.29\% & 29.66\%& 41.52\% & 55.88\% & 20.17\% & 30.40\% & 30.32\% & 46.60\% & 38.33\% & 27.04\% & 44.43\% & 23.47\% & 45.07\% \\
    GPT-4o &LLM& No & 37.38\% & 38.75\% & 43.24\% & 41.50\% & 41.51\% & 57.32\% & 46.87\% & 47.62\% & 46.43\% & 51.56\% & 40.00\% & 37.22\% & 38.15\% & 34.15\% & 41.46\% \\
    InternVL-1.5-Chat-20B &LMM& Yes& 26.78\% & 63.87\% & 54.19\% & 63.57\% & 56.30\% & 41.78\% & 70.81\% & 71.46\% & 70.70\% & 70.95\% & 48.7\% & 51.67\% & 52.77\% & 52.69\% & 55.13\% \\
    GPT-4o &LMM& No& 29.77\% & 63.07\% & 55.22\% & 63.33\% & 56.54\% & 34.45\% & 77.47\% & 75.21\% & 77.20\% & 75.39\% & 39.73\% & 57.32\% & 54.25\%  & 53.63\% & 49.72\% \\
    \midrule
    Human& - & -& 11.11\% & 80.65\% & 69.35\% & 93.56\% & 80.66\% & 12.50\% & 88.12\% & 83.17\% & 89.99\% & 88.77\% & 5.94\% & 94.06\% & 93.07\% & 95.70\% & 95.71\% \\
    \bottomrule
 
\end{tabular}}
    \label{tab:finegrained-exp}
\end{table*}

\subsection{Experiment Results}

As shown in Table~\ref{tab:main-exp}, we compare the performance of different LLMs and LMMs on \dataset and the original benchmarks. We evaluated the performance of graduate students on the benchmarks as a strong baseline. The human evaluation guideline is shown in Appendix~\ref{app:humaneval}. We also include random guess as a weak baseline performance.

We first evaluate LLMs and LMMs separately. For LLMs, all models perform poorly under the Genuine Accuracy metric in \dataset. For example, the advanced GPT-4o achieves $67.62\%$ accuracy on the original ScienceQA-Image benchmark but drops to $10.30\%$ on its calibrated version in \dataset, a decrease of $57.32$ percentage points. Similar declines are observed in MMMU (down $37.38\%$) and MathVista (down $40\%$). These low scores more accurately reflect LLMs' general multimodal capabilities due to their lack of visual perception abilities. Open-source LLMs also show drastic declines in Genuine Accuracy. The Qwen-1.5 series suggests that larger LLMs perform better on both benchmarks, supporting the idea that stronger LLMs are better at guessing. However, this trend does not hold for the LLaMA-3 series, indicating a need for our further investigation in the future. Generally, most LLMs score below $10\%$ in Genuine Accuracy, highlighting the difficulty of the proposed benchmark for LLMs.

On the LMMs side, we also witness a large performance gap between the original benchmarks and \dataset. For proprietary models, GPT-4o and Qwen-VL-Max perform the best. If we compare the performance gap of best LLM and best LMM on the original benchmark and \dataset, we could find that the performance difference in scales is more clear on \dataset. For example, in original MMMU benchmark, GPT-4o with vision ($54.85\%$) is only 1.1 times its non-vision LLM version ($46.09\%$). While in \dataset, the LMM ($25.08\%$) version's performance is $2.9$ times the LLM ($8.71\%$) version of GPT-4o. A similar result also goes for other tasks. The enlarged performance discrepancy is more intuitive given the fundamental functionality difference between LLMs and LMMs. \dataset could better reflect the true abilities of examinees.

\begin{figure*}[!t]
\centering
\includegraphics[width=1\textwidth]{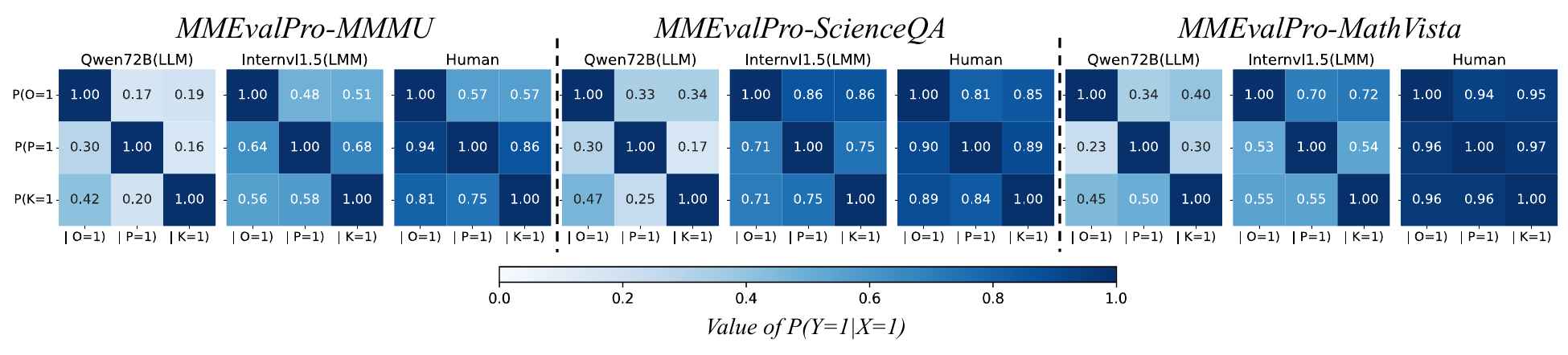}
\caption{Heatmaps of conditional accuracy of \dataset.}
\label{fig:heatmap}

\end{figure*}

\subsection{Fine-grained Analysis}

\begin{figure*}[h]
\includegraphics[width=\textwidth]{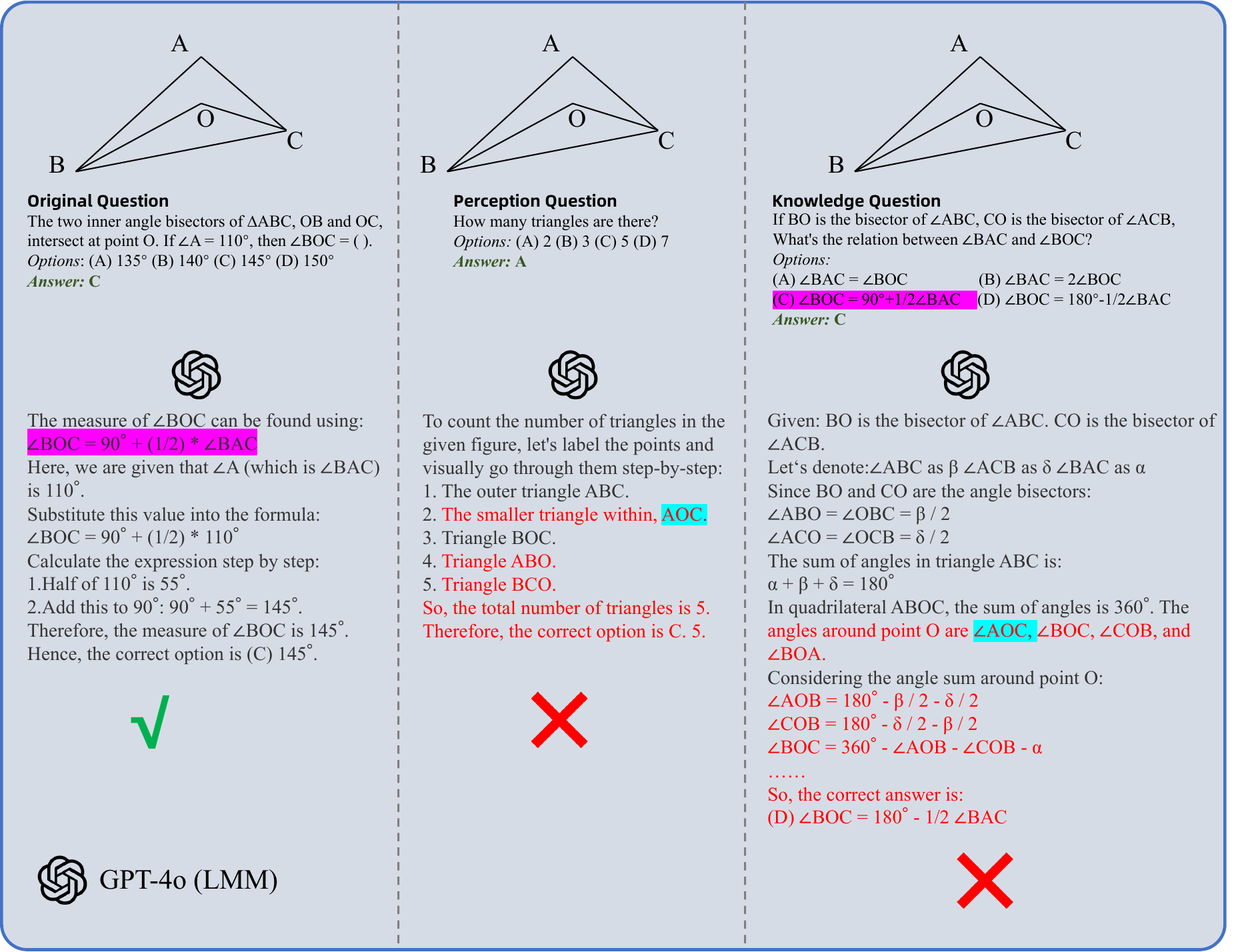}
\caption{Case study on the answer in-consistency problem of LMMs. We could tell from the figure that GPT-4o could answer the original question however failed on the perception and knowledge question. The wrong reasoning process is marked {\color{red}{red}}. The multiply mentioned items are marked with the same background color. In the reasoning process of original question, GPT-4o direcly points out that $\angle$BOC = 90° + (1/2) $\angle$BAC. However in the knowledge question while the "$\angle$BOC = 90° + (1/2) $\angle$BAC" is just one of the options, the model failed to figure it out, while human could easily achieve it. It shows the answer in-consistency problem of the LMM, which can not be pinpointed by single MCQ evaluation. If we further look at the reasoning process of the perception question, we could find that the model could not even recognize the correct number of triangles in the figure. For example it recognized AOC as a trangle, which is actually not. This also causes its problematic reasoning for the knowledge question. }
\label{fig:casestudy1}
\end{figure*}

To better explain models' performance on \dataset, we conducted a fine-grained analysis on the experiments' result according to the metrics proposed in section~\ref{sec:metric}. We select the best performing open-source and proprietary LLMs and LMMs to analyze. The results are shown in Table~\ref{tab:finegrained-exp}. A detailed case analysis is shown in Figure~\ref{fig:casestudy1}. More comparisons of different models are listed in Figure~\ref{fig:cases} from the Appendix.

\paragraph{Why \dataset is challenging and more trustworthy?}

When comparing human performance with that of LLMs and LMMs, we observe a more significant gap in \dataset evaluations than in original benchmarks. This indicates that \dataset is inherently a more challenging task. The primary difficulty stems from the issue of answer consistency, which demonstrates whether the model genuinely understands how to leverage perceptual abilities and knowledge to solve a given problem. To illustrate this, we compare the Consistency Gap (CG) scores among humans, the best-performing LLMs and LMMs. The results suggest that LLMs generally exhibit a larger Consistency Gap than LMMs, while human experts display a considerably smaller CG. This trend is consistent across both open-source and proprietary models. A large Consistency Gap indicates that a model's robustness and generalization abilities are limited: it may be able to answer the original question but fails to respond accurately to related prerequisite questions based on the same image. 

This weakness is difficult to capture for the original benchmarks due to the single MCQ format.  In fact, if we only consider the Average Accuracy of \dataset, as shown in Table~\ref{tab:main-exp}, we observe that the performance difference compared to the original benchmarks is much smaller than the difference in Genuine Accuracy (GA). This suggests that LLMs can often guess correct answers for questions even without image input. \dataset addresses this issue effectively through the GA metric, making it a more reliable evaluation method compared to previous benchmarks.

\paragraph{Why the Consistency Gap is large?}

The gap between Genuine Accuracy and Average Accuracy on the original benchmarks reveals the answer inconsistency problem. We are further interested in what causes the problem. There are two possible reasons for a large Consistency Gap, that the model correctly answer the original question however fails on the perception or knowledge one. We compare the \textbf{Perception Consistency (PC)} and \textbf{Knowledge Consistency (KC)} of the evaluated models and humans. We find that there is a clear performance border between humans and LMMs, LMMs and LLMs. Humans could reach at least $90\%$PC and $80\%$KC in various sub-tasks, showing strong answer consistency. While the numbers for LMMs are $50\%$PC and $55\%$KC, for LLMs are $23\%$PC and $41\%$KC according to Table~\ref{tab:finegrained-exp}. 

PC and KC intuitively reflect the model's likelihood of correctly answering perception and knowledge questions if it has already solved the original question. Low PC and KC scores lead to a significant consistency gap. Beyond PC and KC, we visualize all conditional probabilities for the best open-source LLM and LMM in Figure~\ref{fig:heatmap}. Humans generally exhibit high probabilities of correctly answering one question given a correct answer to another, indicating consistent thinking. In contrast, LLMs show the lowest probabilities compared to LMMs and humans. This is expected, as LLMs lack consistent multimodal problem-solving paths due to their absence of visual perception, thus supporting the credibility of the benchmarks.

Examining the Perception Accuracy (PA) and Knowledge Accuracy (KA) in Table~\ref{tab:finegrained-exp}, we find that LMMs demonstrate a greater advantage in PA compared to KA when contrasted with LLMs. This is because PA depends directly on visual capabilities, which LLMs lack. The above conclusion explains why tested models have larger CG compared to humans, which is a potential and promising direction for future LMMs to improve on.


\section{Related Work}

There have been several benchmarks built for evaluating LMMs~\citep{feng2024bioactivity, yang2024poisoning}, such as MMBench, MME, Seed-Bench~\citep{liu2023mmbench,fu2023mme, li2023seedbench} that assess LMMs performance from multiple fine-grained dimensions.  LVLM-eHub, M3IT~\citep{xu2023lvlmehub,li2023m3it} focus on the general instruction following ability. MMMU, MathVista, ScienceQA~\citep{yue2023mmmu, lu2024mathvista,lu2022scienceqa} require perception from the vision part and knowledge in the language part.

Nonetheless, critiques have been raised regarding the limitations of these existing benchmarks in effectively evaluating LMMs. PCA-Bench, MathVerse~\citep{chen2024pcabench,zhang2024mathverse} adopt strong LLMs such as GPT4 and GPT4-Vision to score the reasoning process of LMMs in embodied-AI and math diagram questions, in order to pinpoint cases where the LMM gets the correct answer by a fluke. Yet, using a proprietary model to conduct evaluation hinders the broader usage of the method, moreover, the evaluation result has bias itself due to the proxy model and would change over time. MMStar~\citep{mmstar} filters out the questions that do not rely on visual information in existing multimodal benchmarks. However, it does not address the issue inherent in MCQ, where models can potentially get the correct answer without truly understanding the content. Compared with those benchmarks, \dataset is more economical, easy-to-use, and calibrated for evaluating multimodal models.

\section{Conclusion}

We propose \dataset, a multimodal benchmark designed to address issues identified in previous evaluations and built upon MMMU, ScienceQA-Image, and MathVista. \dataset introduces twin perception and knowledge anchors to the original framework and defines Genuine Accuracy as its primary metric, thereby reducing the likelihood of LLMs manipulating the questions. Our extensive experiments and analyses on a wide array of models and human experts demonstrate that \dataset more accurately reflects the true capabilities of the tested LMMs and presents a more challenging task. Notably, even the most advanced models, such as GPT-4o and Qwen-VL-Max, trail behind human performance by a substantial gap of more than $30\%$ in Genuine Accuracy. Our analysis into the reasons behind the consistency gap problem elucidates the disparity and provides valuable insights for future research. 

\section*{Limitations}

In order to ensure the high quality and accuracy of \dataset, we employed manual annotation with human experts to construct the dataset. A certain level of human effort and expertise are necessary. To some extent, this requirement may limit the expansion efficiency of \dataset. 

\section*{Acknowledgments}

This paper is partially supported by the National Key Research and Development Program of China with Grant No.2023YFC3341203 as well as the National Natural Science
Foundation of China with Grant Numbers 61876004 and 62306014.
\bibliography{custom}

\newpage

\appendix

\section*{Appendix}


\section{Details on \dataset}

\subsection{Data Source}
\label{app:data source}

\textbf{MMMU~\citep{yue2023mmmu}:} The MMMU is a benchmark designed to evaluate multimodal models on massive multi-discipline tasks. 
The benchmark sources its questions from college examinations, quizzes, and textbooks, encompassing $30$ subjects and $183$ subfields across six core disciplines, i.e. Art \& Design, Business, Science, Health \& Medicine, Humanities \& Social Science and Tech \& Engineering. 
This benchmark is meticulously designed to assess models’ capabilities in handling multi-disciplinary tasks, drawing upon college-level subject knowledge. 

\textbf{ScienceQA~\citep{lu2022scienceqa}:} The Science Question Answering (ScienceQA) is another pivotal resource, which consists of 21,208 multimodal multiple choice questions with diverse science topics.
There are only $48.7\%$ questions of ScienceQA that have an image context.
It is renowned for its application in multimodal tasks and features a domain diversity spanning three primary science subjects, i.e., natural, language, and social.
The dataset comprises multimodal science questions that are collated from elementary and high school science curricula, ensuring a breadth of scientific inquiry and comprehension. 
The subjects of the questions can be categorized by Biology, Physics, Chemistry, and others.

\textbf{MathVista~\citep{lu2024mathvista}:} The MathVista benchmark is developed to evaluate the reasoning ability of the multimodal models, which consists of $6,141$ examples from $31$ datasets ($28$ mathematics and IQTest, FunctionQA, PaperQA).
The dataset offers exclusively mathematical and visual tasks. 
This source enriches our dataset with rigorous computational and analytical problems, providing a robust framework for evaluating quantitative reasoning in multimodal contexts.

Finally, we select the validation set of MMMU ($722$ questions), the questions with images in ScienceQA ($2,097$ questions), and the testmini set of MathVista ($540$ questions) to construct the \dataset.
From the $722$ questions in the validation set of MMMU, we chose the questions with topics suited for the annotator's major and other questions in easy-level to annotate, which resulted in $339$ final questions. And we annotated all questions selected from the testmini set of MathVista and $1,259$ of ScienceQA-Image.

The distribution of \dataset is shown in the Figure~\ref{fig:dataset-dist}.
There are $58.89\%$ questions originating from ScienceQA, $25.26\% $from MathVista, and $15.86\%$ from MMMU.
We further categorize the problems by task categories, subjects, and domains, computing their relative percentages to the total questions.
In the annotated questions originated from ScienceQA, there are $16.68\%$ of them subject to Geography, $12.61\%$ subject to Physics, $12.02\%$ with the Biology topic, and the remaining questions distributed in History, Chemistry, Economics, etc.
In MathVista, the annotated questions contain 5 tasks, including GPS(geometry problem solving), VQA(visual question answering), FQA(figure question answering), TQA(textbook question answering), and MWP(math word problem), the last one only accounted for a very small proportion and is not annotated in the figure.
According to the subjects of questions from MMMU, there are $4.91\%$ questions in Medicine, $3.27\%$ in Art, $3.65\%$ in Business, $2.71\%$ in Science, and $1.32\%$ from other subjects with easy-level.

\subsection{License}

We check all the datasets' licenses and they all permit customization and redistribution for non-commercial use. MMMU is under Apache 2.0 License, ScienceQA is under MIT License and MathVista is covered by CC BY-SA 4.0. 

\dataset can be used commercially as a test set, but using it as a training set is prohibited. By accessing or using this dataset, users acknowledge and agree to abide by these terms in conjunction with the CC BY-SA 4.0 license. We make sure there is no offensive content found during the whole annotation process.

\subsection{\dataset Examples}

\begin{figure}[!h]
\centering
\includegraphics[width=0.5\textwidth]{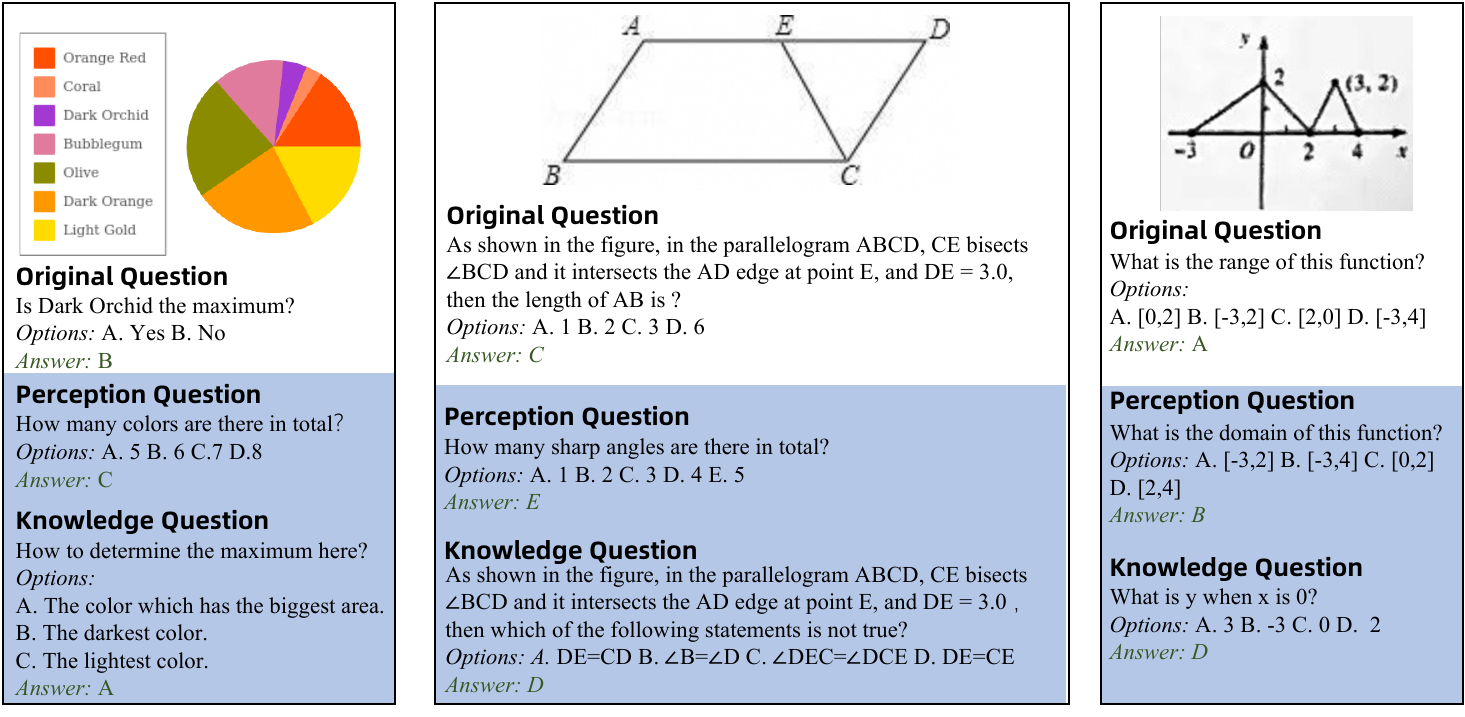}
\caption{Examples of the MathVista subset of \dataset.}
\label{fig:examples-mathvista}
\end{figure}

\begin{figure}[!h]
\centering
\includegraphics[width=0.5\textwidth]{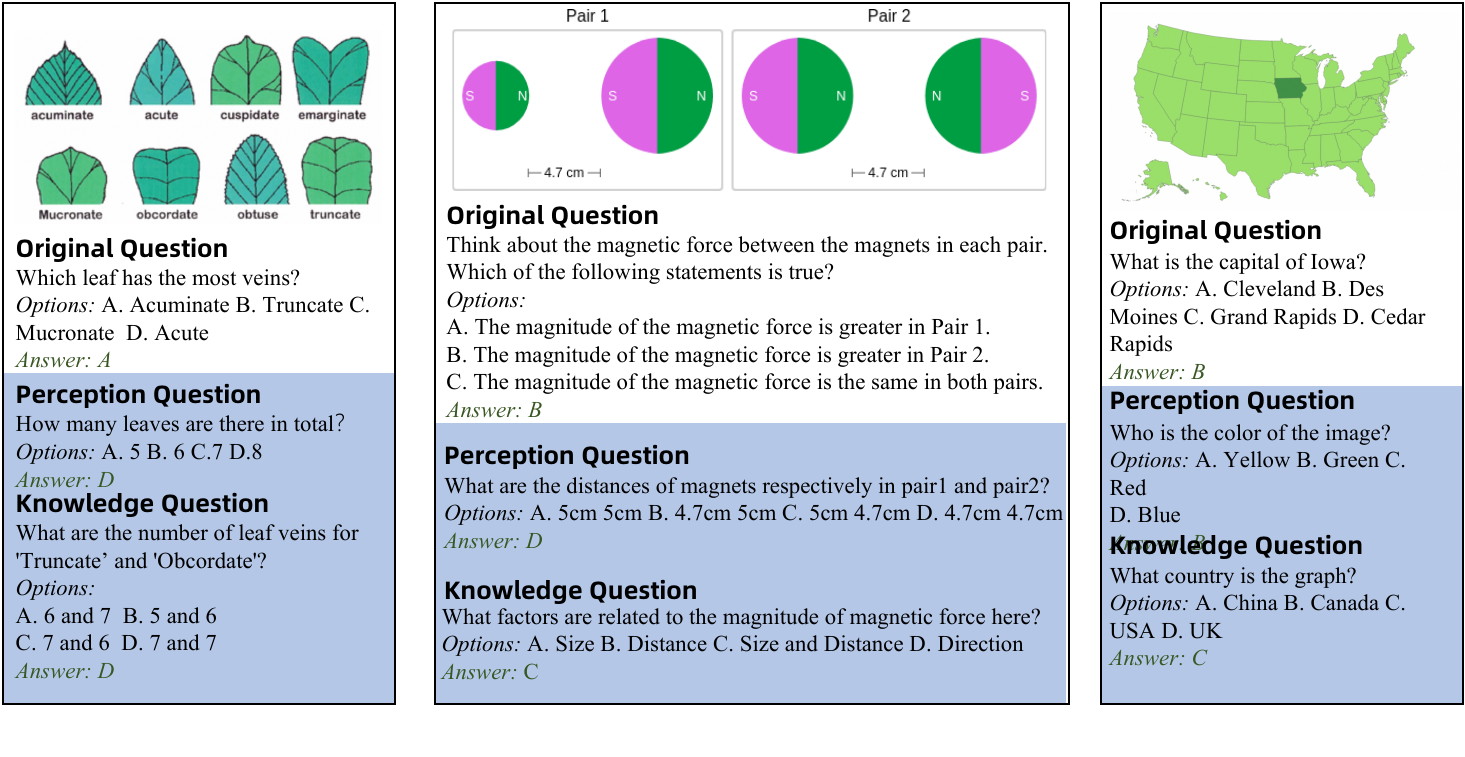}
\caption{Examples of the ScienceQA subset of \dataset.}
\label{fig:examples-Sciqa}
\end{figure}

\begin{figure}[!h]
\centering
\includegraphics[width=0.5\textwidth]{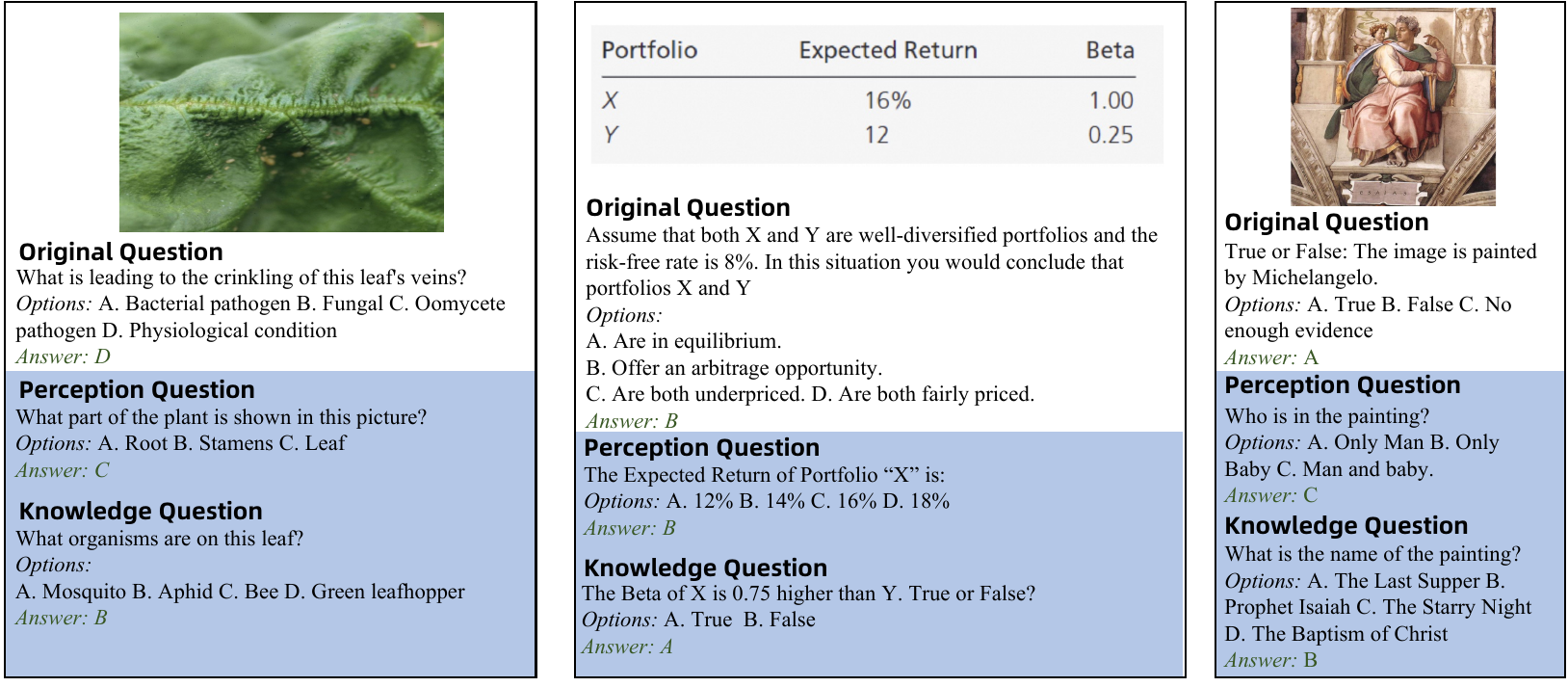}
\caption{Examples of the MMMU subset of \dataset.}
\label{fig:examples-mmmu}
\end{figure}

We list examples of different splits of \dataset as shown in Figure~\ref{fig:examples-mathvista}, \ref{fig:examples-Sciqa} and \ref{fig:examples-mmmu}.


\subsection{Statistics of \dataset}

\begin{figure}[t]
  \includegraphics[width=0.9\linewidth]{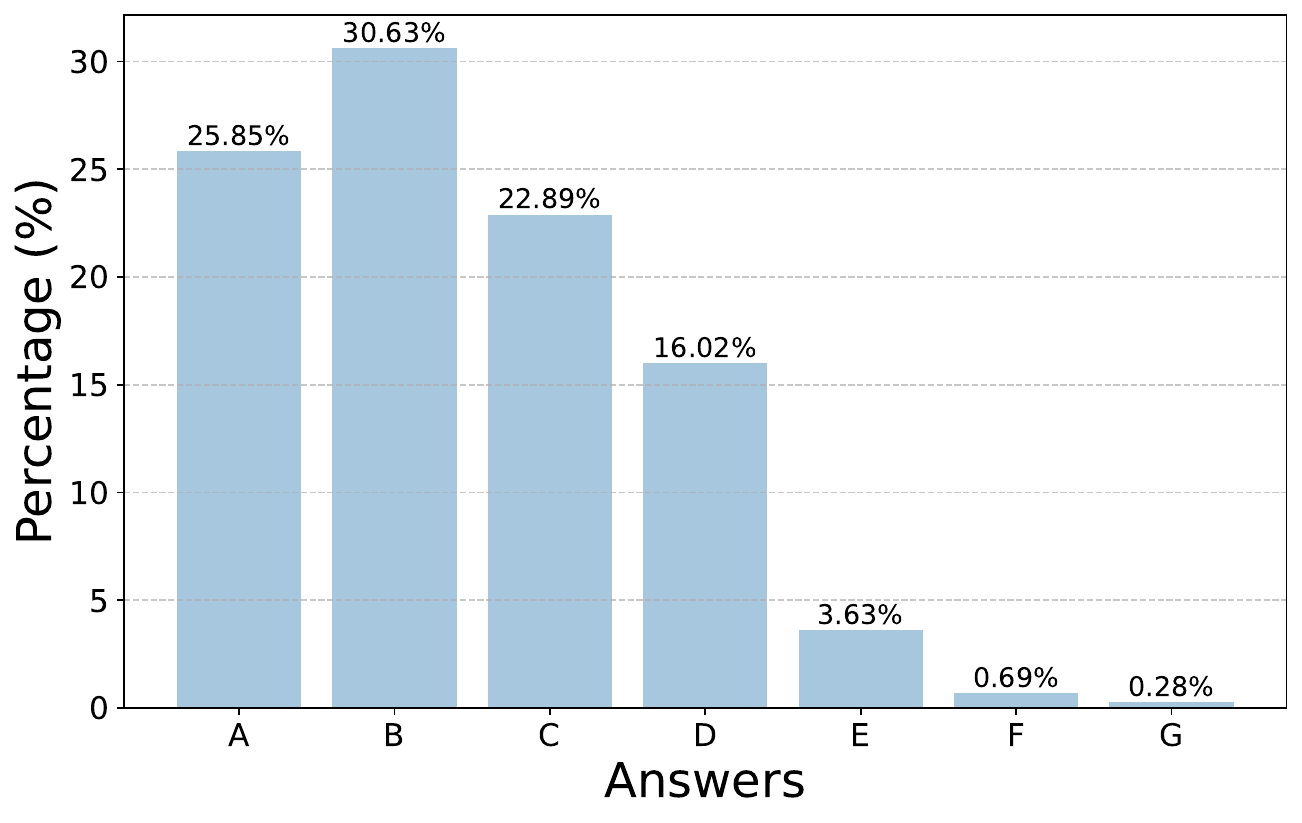} \hfill
  \includegraphics[width=0.9\linewidth]{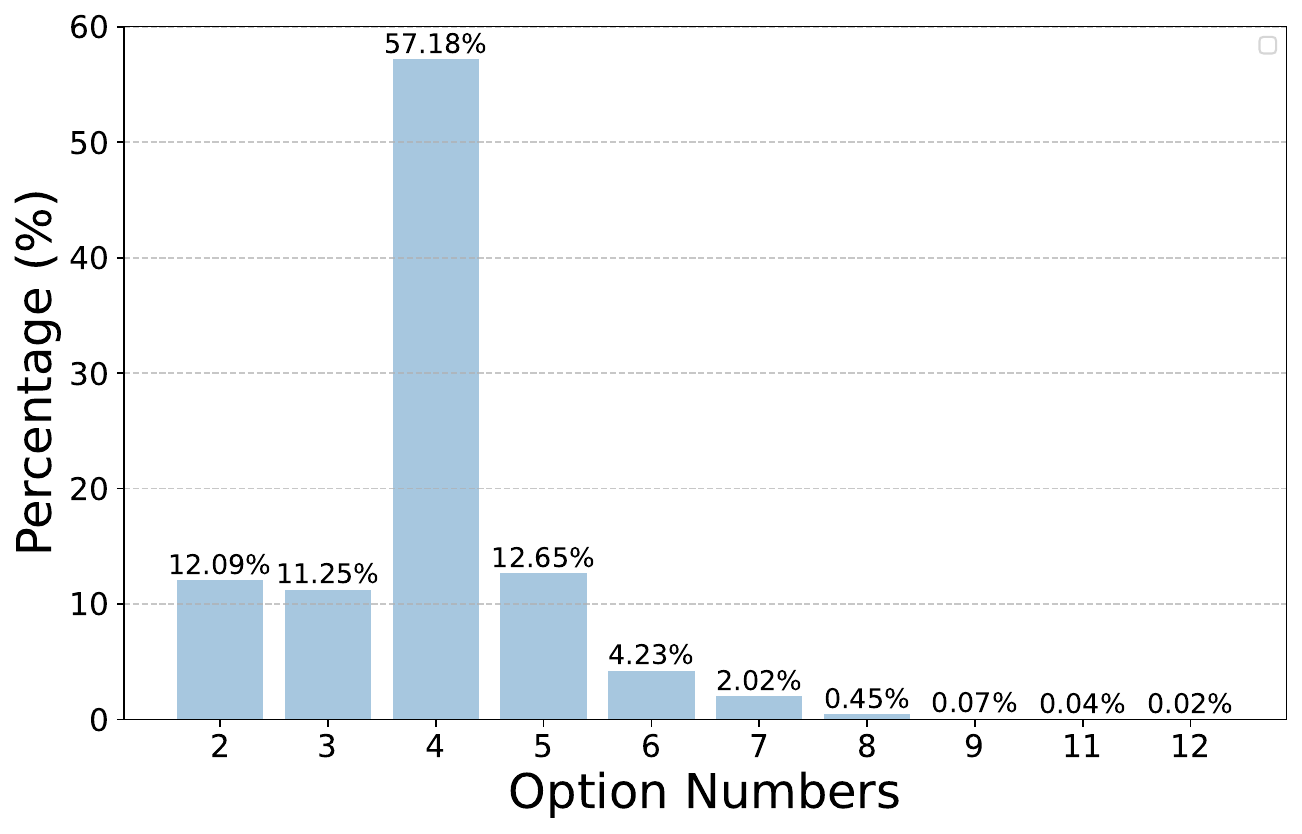}
  \vspace{0.1cm}
  \captionof{figure}{Distribution of Answer Choices in \dataset.}
  \label{fig:dis of answer}
\end{figure}

The key statistics of \dataset is shown in the table~\ref{tab:statistics}. Figure~\ref{fig:dis of answer} display the distribution of answers and the number of options in the \dataset.

\begin{table}[!t]
\centering
\resizebox{0.7\columnwidth}{!}{
\begin{tabular}{lc}
 \toprule
 {\textbf{Statistic}} & {\textbf{Number}} \\
 \midrule
  Source datasets & 3 \\
  Number of question triplets & 2,138 \\
  Number of unique questions & 6,414 \\
  Triplets from MMMU & 339 \\
  Triplets from ScienceQA & 1259 \\
  Triplets from MathVista & 540 \\
 \midrule
 Maximum question length & 165 \\
 Maximum choice number & 12\\
 Average question length & 9.60 \\
 Average choice number & 3.94 \\
 \bottomrule
 \end{tabular}}
 \captionof{table}{Key statistics of \dataset.}
 \label{tab:statistics}
\end{table}

\newpage
\section{Experiment Setup}
\label{app: Experiment Setup}

\subsection{Prompt Format for Different Models}

\paragraph{For LLM} "Given a question you need to choose the best answer from the given options. I will first give you an example, you need to follow the output format of the answer. Example: Question: \{example question\} Options: \{example options\} Output: \{example answer\}. Just answer the following question with only the letter of the correct option, or you will get no credit. Question:\{question\} Options: \{options\}"

We use a fixed demonstration for all inferences to format the output. The example is:"\textbf{Question:} Baxter Company has a relevant range of production between 15,000 and 30,000 units. The following cost data represents the average variable costs per unit for 25,000 units of production. If 30,000 units are produced, what are the per unit manufacturing overhead costs incurred? \textbf{Options:} (A) \$6 (B) \$7 (C) \$8 (D) \$9 \textbf{Output:} A".

\paragraph{For LMM} "Analyse the image and choose the best answer for the following question:\{question\} \\\ Options: \{options\} Just output the letter of the correct answer."

\subsection{Model Hyper-parameters}

For open-source models, we use the default inference script provided in corresponding papers and githubs. For proprietary models, we follow the official guide to call the API. In particular, we do not use sampling techniques during generation to ensure our results are reproducible. All experiments are done on a local server with 4 NVIDIA-A100 GPUs.

\newpage
\section{\dataset Triplet Annotation Guideline}
\label{app:annotation-guideline}

\begin{figure}[!h]
\centering
\includegraphics[width=0.5\textwidth]{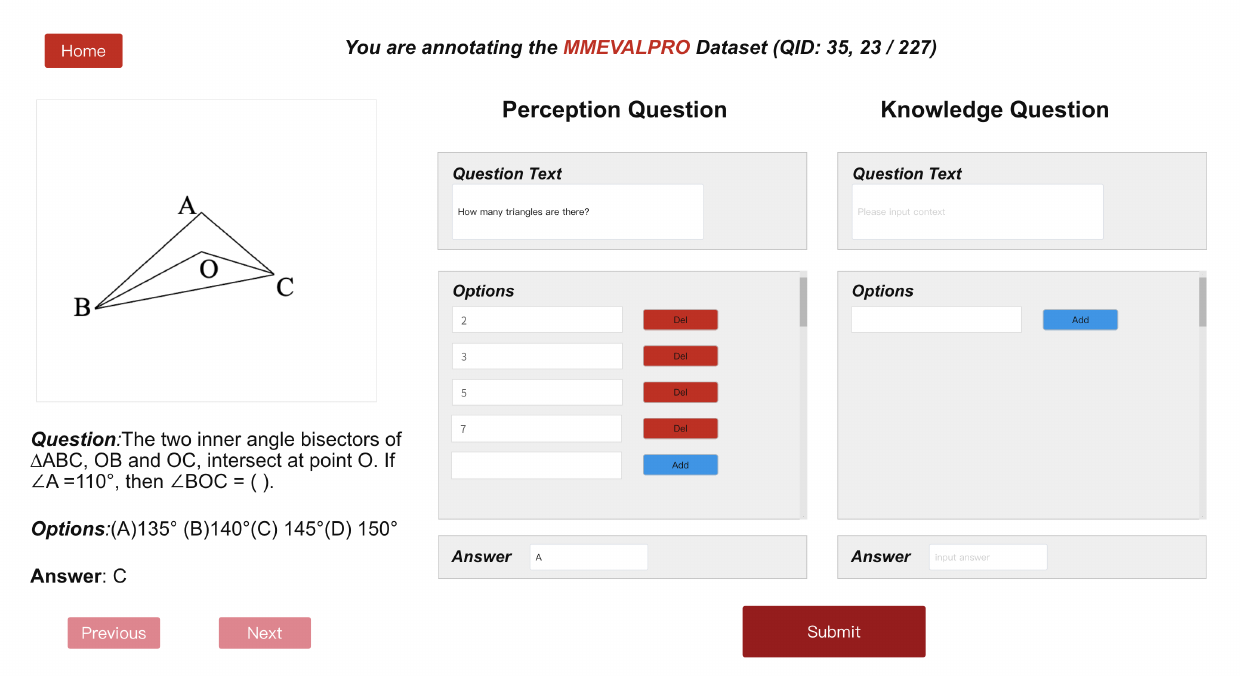}
\caption{UI for annotations of \dataset.}
\label{fig:human annotation}
\end{figure}

\begin{figure}[!h]
\centering
\includegraphics[width=0.5\textwidth]{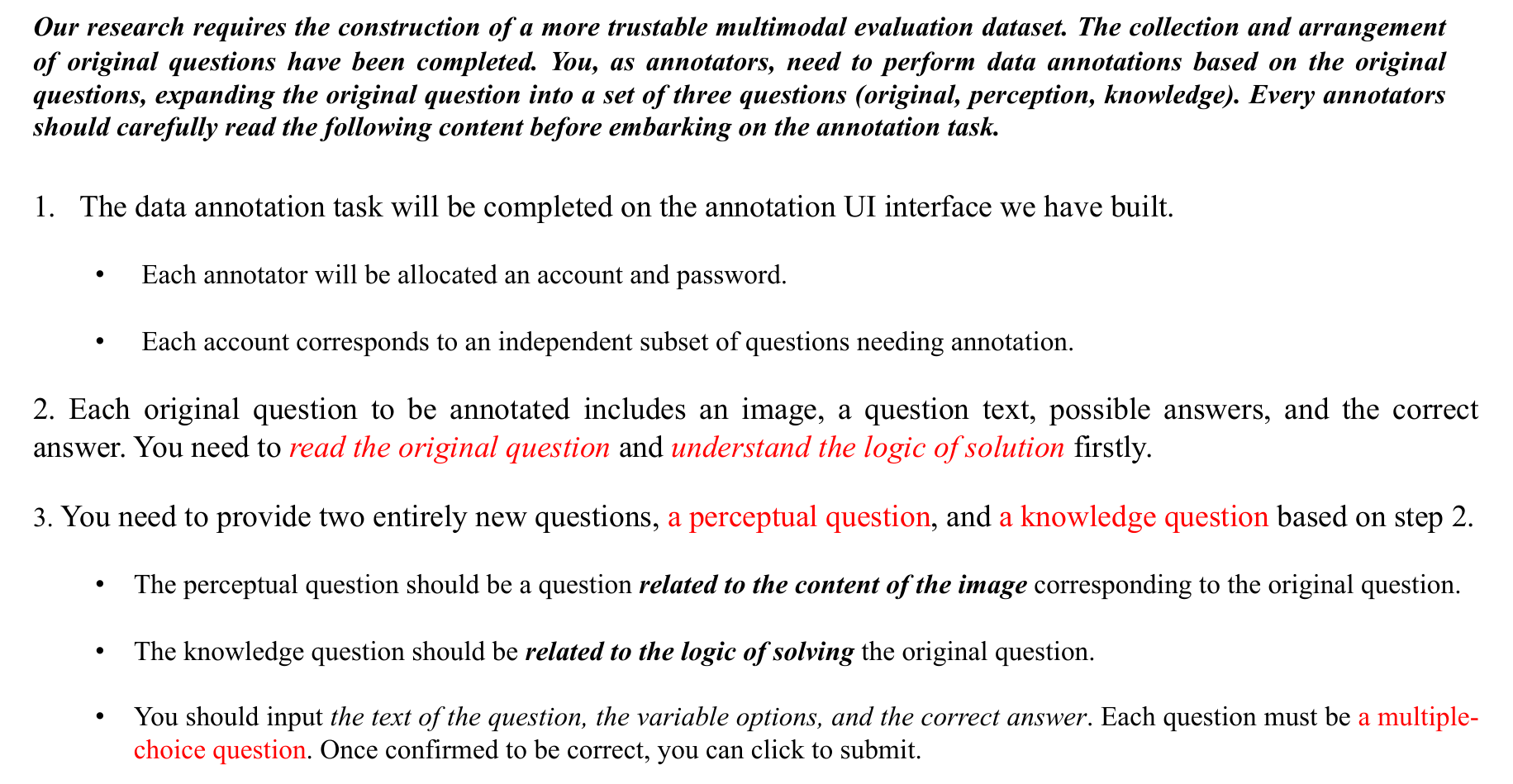}
\caption{Annotation Guideline of \dataset.}
\label{fig:human annotation guide}
\end{figure}

We developed an annotation Web UI to enable expert annotators to construct question triplets of \dataset.
The Web UI is shown in the figure~\ref{fig:human annotation}.
The annotators of \dataset were trained with the guideline shown in figure~\ref{fig:human annotation guide} before formal work.

\newpage
\section{\dataset Triplet Checking Guideline}
\label{app:check-guideline}
\begin{figure}[!h]
\centering
\includegraphics[width=0.5\textwidth]{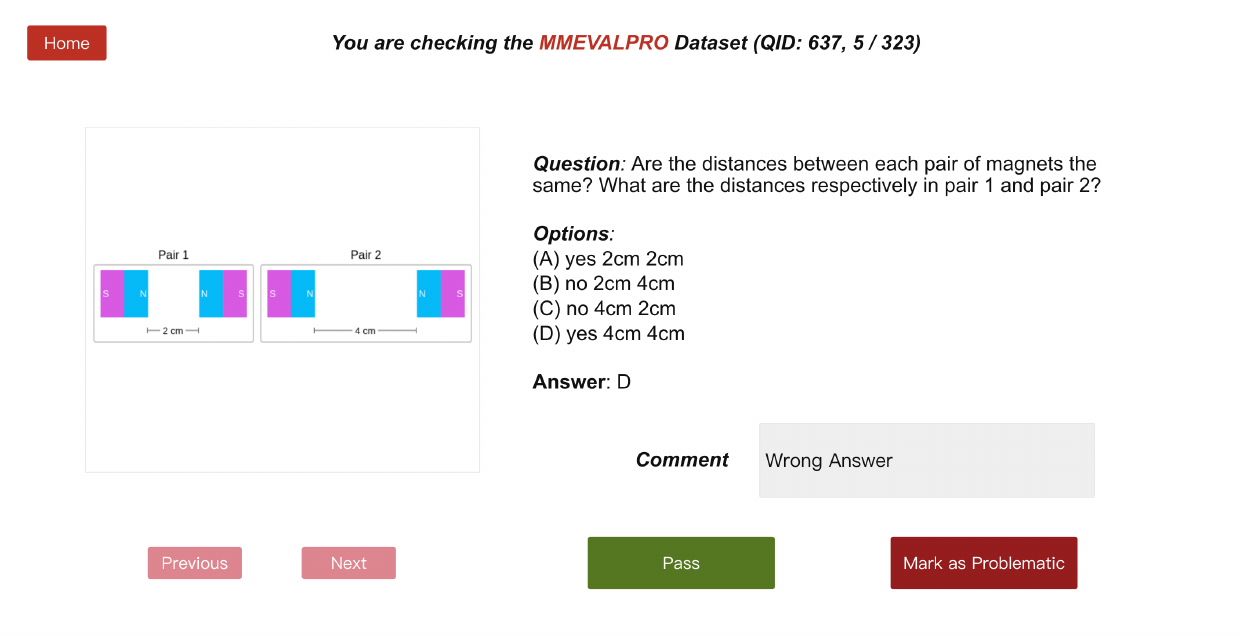}
\caption{UI for checking of \dataset.}
\label{fig:human check}
\end{figure}

\begin{figure}[!h]

\centering
\includegraphics[width=0.5\textwidth]{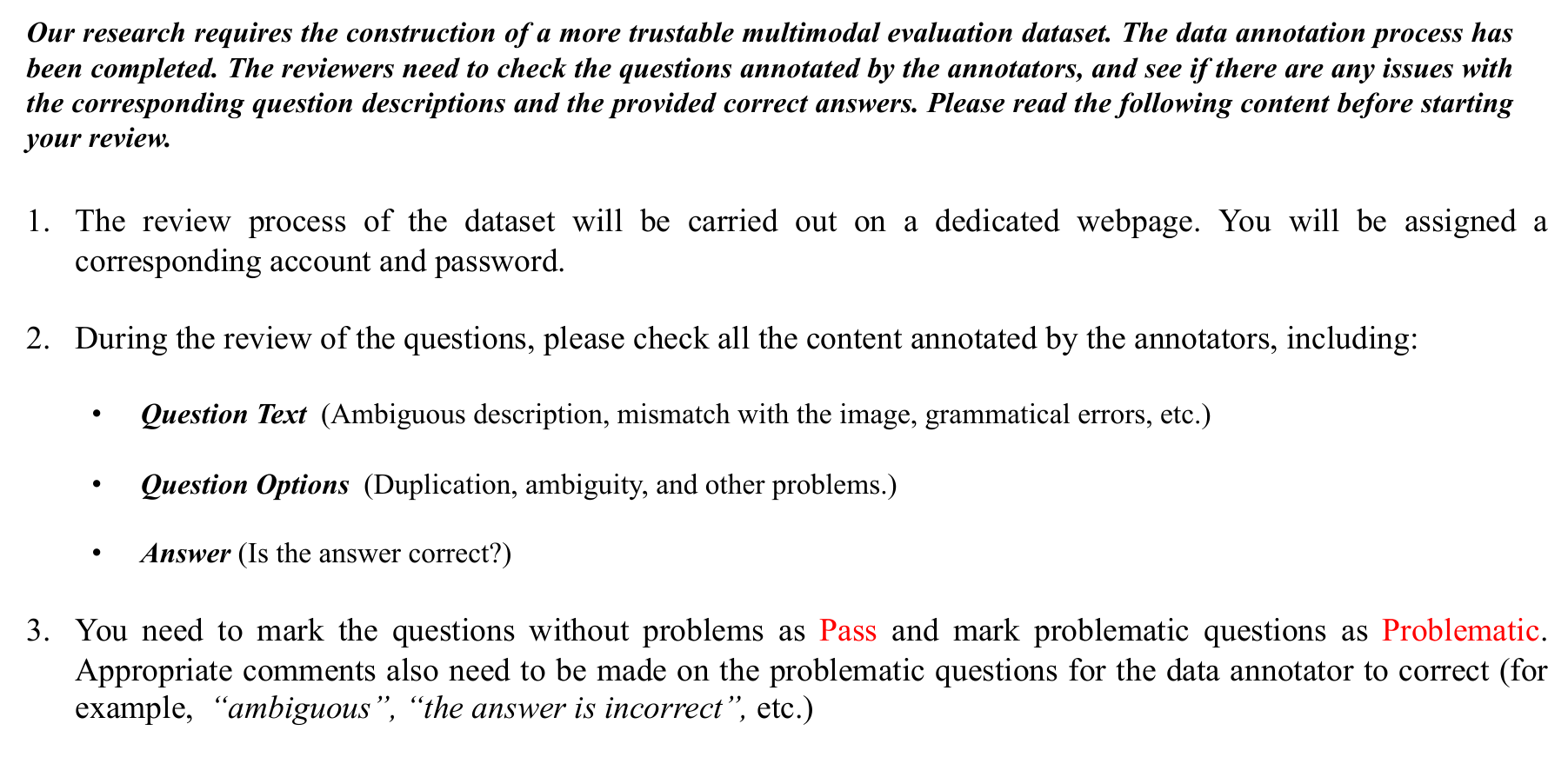}
\caption{Checking Guideline of \dataset.}
\label{fig:human check guide}
\end{figure}

In our study, we employed double checking to maintain the quality of \dataset.
We also developed a web page shown in figure~\ref{fig:human check} for checking.
The checkers were trained with checking instructions shown in figure~\ref{fig:human check guide}.

\newpage
\section{Human Evaluation Guide}
\label{app:humaneval}

\begin{figure}[!h]
\centering
\includegraphics[width=0.5\textwidth]{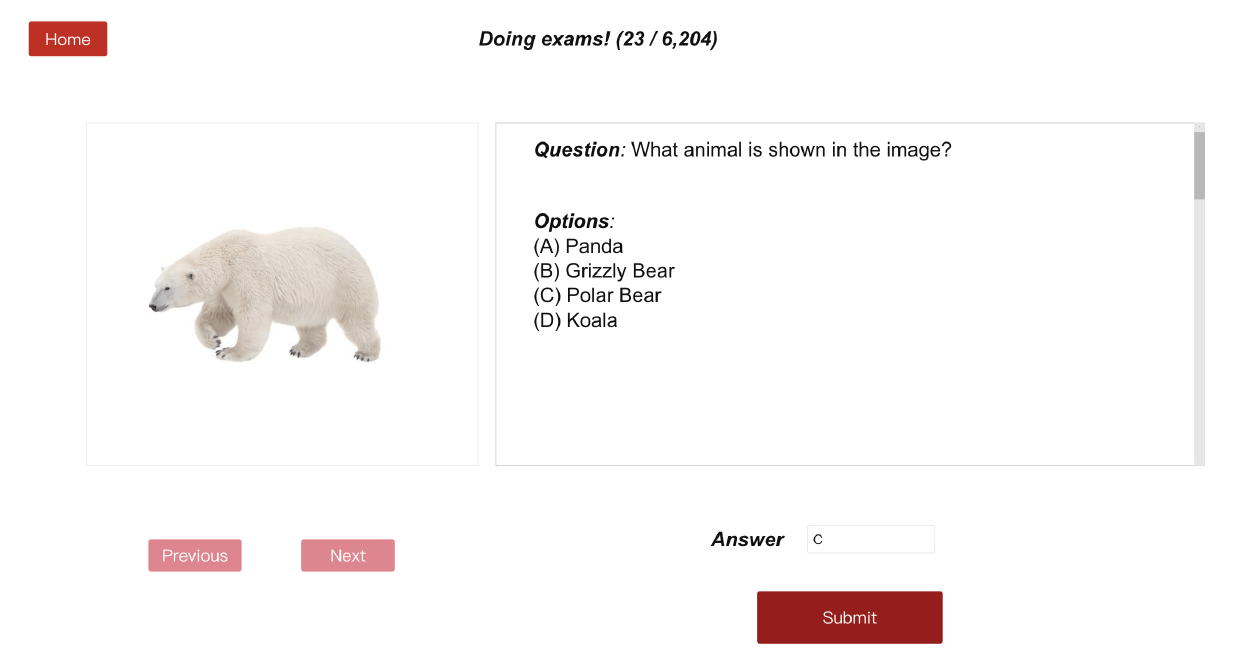}
\caption{UI for human evaluation of \dataset.}
\label{fig:human eval}
\end{figure}

To ensure the comprehensiveness of our study, we employed five graduates with special knowledge to do human evaluation in \dataset.
The designed web ui of human evaluation is shown in figure~\ref{fig:human eval}.

\begin{figure*}[!h]
\centering
\includegraphics[width=0.9\textwidth]{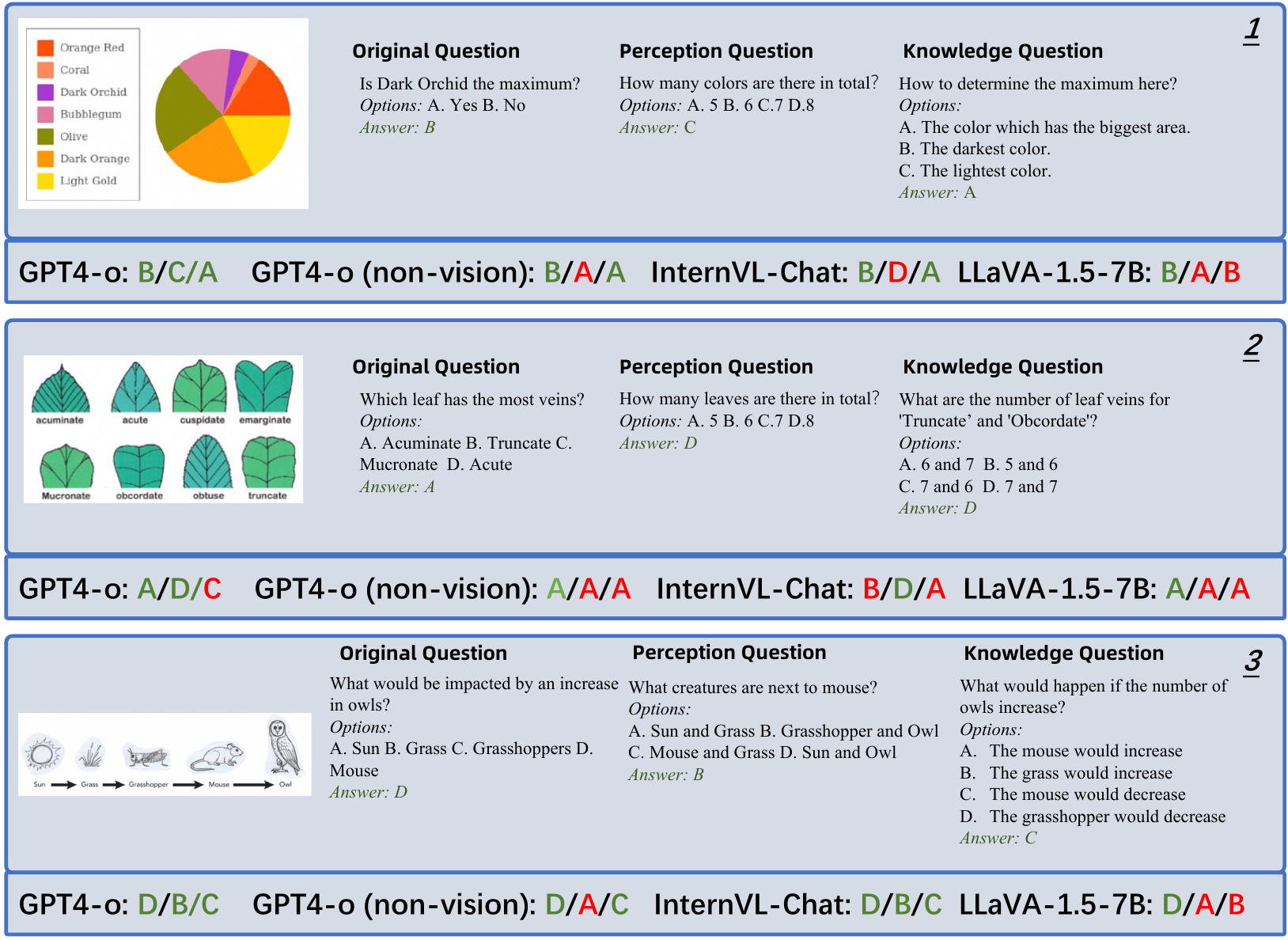}
\caption{Cases of different models' comparison for \dataset.}
\label{fig:cases}
\end{figure*}

\section{Comparison of Different Models}

We list three cases comparing the output of GPT-4o~\citep{gpt4o}, GPT-4o (non-vision)~\citep{gpt4o}, InternVL-Chat~\citep{chen2023internvl} and LLaVA-1.5~\citep{liu2023llava}. We observed that, while both GPT-4o (non-vision) and LLaVA-1.5 could answer all the origin questions correctly, they struggled with most perception and knowledge questions. This highlights a Type-I error in current evaluation benchmarks: correctly answering a question does not necessarily indicate genuine understanding by the model. On the other hand, more advanced models like GPT-4o and InternVL-Chat demonstrate higher consistency in answering different types of questions.


\centering

\end{document}